\documentclass{llncs}
\usepackage{capt-of}
\usepackage{booktabs,caption,fixltx2e}
\usepackage[flushleft]{threeparttable}
\usepackage{graphicx}
\usepackage{makeidx} 
\usepackage{amsmath}
\usepackage{multirow}

\usepackage{amsfonts}%
\usepackage{amssymb}%
\usepackage[english,oldcommands,plain,lined]{algorithm2e}
\usepackage{graphics}
\usepackage{subfig}
\usepackage[labelfont=bf]{caption}
\usepackage[font={small}]{caption}
\usepackage{transparent}
\usepackage{color}
\usepackage{breqn}
\usepackage{epstopdf}
\usepackage{wrapfig}

\usepackage{color, colortbl}
\definecolor{LightCyan}{rgb}{0.88,1,1}
 \usepackage{cite}
\usepackage[font=small,labelfont=bf]{caption}
\usepackage{enumitem}
\usepackage{multirow}
\usepackage{hyperref}
\newcolumntype{P}[1]{>{\centering\arraybackslash}p{#1}}

\begin{document}

\title{Concurrent Spatial and Channel `Squeeze \& Excitation' in Fully Convolutional Networks}

\author{Abhijit Guha Roy\inst{1,2}, Nassir Navab\inst{2,3}, Christian Wachinger\inst{1}}
\institute{
$^1$Artificial Intelligence in Medical Imaging (AI-Med), KJP, LMU M\"{u}nchen, Germany.\\
$^2$Computer Aided Medical Procedures, Technische Universit\"{a}t M\"{u}nchen, Germany.\\
$^3$Computer Aided Medical Procedures, Johns Hopkins University, USA.
}

\maketitle 
\begin{abstract}
Fully convolutional neural networks (F-CNNs) have set the state-of-the-art in image segmentation for a plethora of applications. Architectural innovations within F-CNNs have mainly  focused on improving spatial encoding or network connectivity to aid gradient flow. In this paper, we explore an alternate direction of recalibrating the feature maps adaptively, to  boost meaningful features, while suppressing weak ones. We draw inspiration from the recently proposed squeeze \& excitation (SE)  module for channel recalibration of  feature maps for  image classification. Towards this end, we introduce three variants of SE modules for image segmentation, (i) squeezing spatially and exciting channel-wise (cSE), (ii) squeezing channel-wise and exciting spatially (sSE) and (iii) concurrent spatial and channel squeeze \& excitation (scSE). We effectively incorporate these SE modules within three different state-of-the-art F-CNNs (DenseNet, SD-Net, U-Net) and observe consistent improvement of performance across all architectures, while minimally effecting model complexity. Evaluations are performed on two challenging applications: whole brain segmentation on MRI scans and organ segmentation on whole body contrast enhanced CT scans. 

\end{abstract}

\section{Introduction}
\label{sec:intro}
Deep learning, in particular, convolutional neural networks (CNN) have become the standard for image classification~\cite{alexnet2012,resnet}. Fully convolutional neural networks (F-CNNs) have become the tool of choice for many image segmentation tasks in medical imaging~\cite{Unet,ecb2017,quickNat2018} and computer vision~\cite{longfcn2015,deconvnet2015,segnet,densenet}. The basic building block for all these architectures is the convolution layer, which learns filters capturing local spatial pattern along all the input channels and generates feature maps jointly encoding the spatial and channel information. While much effort is put into improving this joint encoding of spatial and channel information, encoding of the spatial and channel-wise patterns independently is less explored. Recent work attempted to address this issue by explicitly modeling the interdependencies between the channels of  feature maps. An architectural component called squeeze \& excitation (SE) block~\cite{SE2017} was introduced, which can be seamlessly integrated within any CNN model. The SE block factors out the spatial dependency by global average pooling to learn a channel specific descriptor, which is used to recalibrate the feature map to emphasize on useful channels. Its nomenclature is motivated by the fact that the SE block `squeezes' along the spatial domain and `excites' or reweights along the channels. A convolutional network with SE blocks won the first place in the ILSVRC 2017 classification competition on the ImageNet dataset, indicating its effectiveness~\cite{SE2017}.

In this work, we want to leverage the high performance of SE blocks for image classification to image segmentation with F-CNNs. We refer to the previously introduced SE block as channel SE (cSE), because it  only excites channel-wise, which proved to be effective for classification. We hypothesize that for image segmentation, the pixel-wise spatial information is more informative. Hence, we introduce another SE block, which `squeezes' along the channels and `excites' spatially, termed \emph{spatial} SE (sSE). Finally, we propose to have concurrent spatial and channel SE blocks (scSE) that recalibrate the feature maps separately  along channel and space, and then combines the output. Encouraging feature maps to be more informative both spatially and channel-wise. To the best of our knowledge, this is the first time that spatial squeeze \& excitation is proposed for neural networks  and the first integration of squeeze \& excitation  in F-CNNs. 

We integrate the proposed SE blocks (cSE, sSE and scSE) in three state-of-the-art F-CNN models for image segmentation to demonstrate that SE blocks are a generic network component to boost performance. We evaluate the segmentation performance in two important medical applications: whole-brain and whole-body segmentation. In whole-brain segmentation,  we automatically identify 27 cortical and subcortical structures on magnetic resonance imaging (MRI) T1-weighted brain scans. In whole-body segmentation, we automatically label 10 visceral organs on contrast-enhanced CT scan of the abdomen.

\noindent
\textbf{Related Work: }
F-CNN architectures have successfully been used in a wide variety of medical image segmentation tasks to provide state-of-the-art performance. A seminal F-CNN model is U-Net~\cite{Unet}, which has an encoder/decoder based architecture combined with skip connections between encoder and decoder blocks with similar spatial resolution. SkipDeconv-Net (SD-Net)~\cite{ecb2017} builds upon U-Net, using unpooling layers used in \cite{deconvnet2015} for decoding, learnt by jointly optimizing logistic and Dice loss functions. A more recent architecture introduces dense connectivity within the encoder and decoder blocks, unlike U-Net and SD-Net which uses normal convolutions, termed fully convolutional DenseNet~\cite{densenet}.

\section{Methods}
\begin{figure}[h]
\centering
\includegraphics[width=0.9\textwidth]{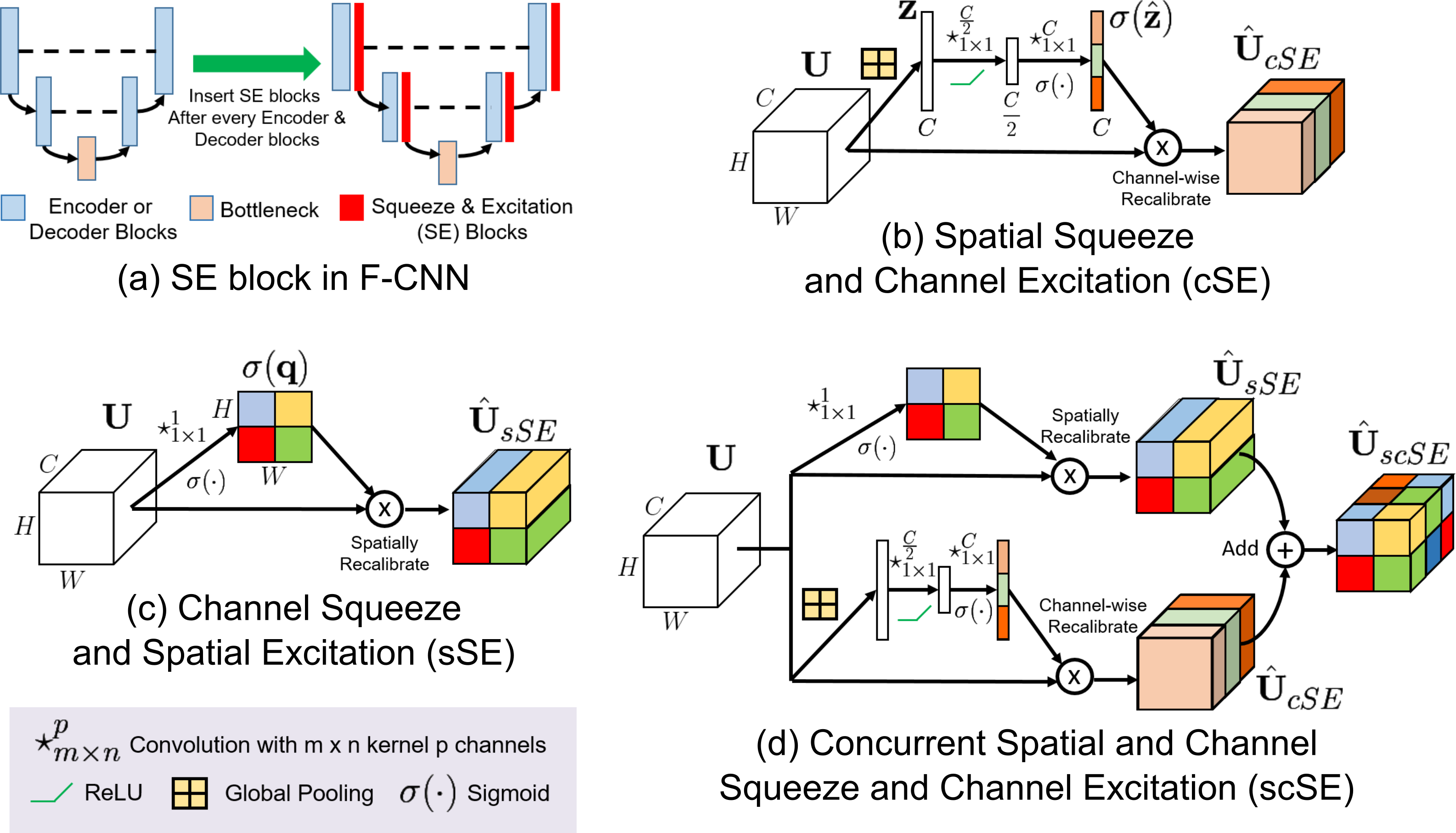}
\vspace{-2mm}
\caption{Illustration of network architecture with squeeze \& excitation (SE) blocks. (a) The proposed integration of SE blocks within F-CNN. (b-d) The architectural design of  cSE, sSE and scSE blocks, respectively, for recalibrating feature map $\mathbf{U}$.
}
\label{fig:GA}
\end{figure}

Let us assume an input feature map $\mathbf{X} \in \mathbb{R}^{H \times W \times C'}$ that passes through an encoder or decoder block $\mathbf{F}_{tr}(\cdot)$ to generate output feature map $\mathbf{U} \in \mathbb{R}^{H \times W \times C}$,  $\mathbf{F}_{tr}:\mathbf{X}\rightarrow\mathbf{U}$.  Here $H$ and $W$ are the spatial height and width, with $C'$ and $C$ being the input and output channels, respectively. The generated $\mathbf{U}$  combines the spatial and channel information of $\mathbf{X}$ through a series of convolutional layers and non-linearities defined by $\mathbf{F}_{tr}(\cdot)$. We place the SE blocks $\mathbf{F}_{SE}(\cdot)$ on $\mathbf{U}$ to recalibrate it to $\hat{\mathbf{U}}$. We propose three different variants of SE blocks, which are detailed next. The SE blocks can be seamlessly integrated within any F-CNN model by placing them after every encoder and decoder block, as illustrated in Fig.~\ref{fig:GA}(a). $\hat{\mathbf{U}}$ is used in the subsequent pooling/upsampling layers.

\noindent
\subsection{Spatial Squeeze and Channel Excitation Block (cSE)}
We describe the spatial squeeze and channel excitation block, which was proposed in~\cite{SE2017}. We consider the input feature map $\mathbf{U} = [\mathbf{u}_1, \mathbf{u}_2, \cdots, \mathbf{u}_{C}]$ as a combination of channels $\mathbf{u}_i \in \mathbb{R}^{H \times W}$. Spatial squeeze is performed by a global average pooling layer, producing vector $\mathbf{z} \in \mathbb{R}^{1 \times 1 \times C}$ with its $k^{th}$ element

\begin{equation}
z_k = \frac{1}{H \times W} \sum_i^H \sum_j^W \mathbf{u}_k (i,j).
\end{equation}

\noindent
This operation embeds the global spatial information in vector $\mathbf{z}$. This vector is transformed to $\hat{\mathbf{z}}=\mathbf{W}_1 (\delta(\mathbf{W}_2 \mathbf{z}))$, with $\mathbf{W}_1 \in \mathbb{R}^{C \times \frac{C}{2}}$, $\mathbf{W}_2 \in \mathbb{R}^{\frac{C}{2} \times C}$ being weights of two fully-connected layers and the ReLU operator $\delta(\cdot)$. This encodes the channel-wise dependencies. The dynamic range of the activations of $\hat{\mathbf{z}}$ are brought to the interval $[0, 1]$, passing it through a sigmoid layer $\sigma(\hat{\mathbf{z}})$. The resultant vector is used to recalibrate or excite $\mathbf{U}$ to

\begin{equation}
\hat{\mathbf{U}}_{cSE} = F_{cSE}(\mathbf{U}) = [\sigma(\hat{z_1})\mathbf{u}_1, \sigma(\hat{z_2})\mathbf{u}_2, \cdots, \sigma(\hat{z_{C}})\mathbf{u}_{C}].
\end{equation}

\noindent
The activation $\sigma(\hat{z}_i)$ indicates the importance of the $i^{th}$ channel, which are rescaled. As the network learns, these activations are adaptively tuned to ignore less important channels and emphasize the important ones. The architecture of the block is illustrated in Fig.~\ref{fig:GA}(b).

\vspace{-2mm}
\subsection{Channel Squeeze and Spatial Excitation Block (sSE)}
We introduce the channel squeeze and spatial excitation block that squeezes the feature map $\mathbf{U}$ along the channels and excites spatially, which we consider important for fine-grained image segmentation. Here, we consider an alternative slicing of the input tensor $\mathbf{U} = [\mathbf{u}^{1,1}, \mathbf{u}^{1,2}, \cdots, \mathbf{u}^{i,j}, \cdots ,\mathbf{u}^{H,W}]$, where $\mathbf{u}^{i,j} \in \mathbb{R}^{1 \times 1 \times C}$ corresponding to the spatial location $(i,j)$ with $i \in \{1, 2, \cdots, H\}$ and $j \in \{1, 2, \cdots, W\}$. The spatial squeeze operation is achieved through a convolution $\mathbf{q} = \mathbf{W}_{sq} \star \mathbf{U}$ with weight $\mathbf{W}_{sq} \in \mathbb{R}^{1 \times 1 \times C \times 1}$, generating a projection tensor $\mathbf{q} \in \mathbb{R}^{H \times W}$. Each $q_{i,j}$ of the projection represents the linearly combined representation for all channels $C$ for a spatial location $(i,j)$. This projection is passed through a sigmoid layer $\sigma(.)$ to rescale activations to $[0,1]$, which is used to recalibrate or excite $\mathbf{U}$ spatially

\begin{equation}
\hat{\mathbf{U}}_{sSE} = F_{sSE}(\mathbf{U}) = [\sigma(q_{1,1})\mathbf{u}^{1,1}, \cdots, \sigma(q_{i,j})\mathbf{u}^{i,j}, \cdots, \sigma(q_{H,W})\mathbf{u}^{H,W}].
\end{equation}

\noindent
Each value $\sigma(q_{i,j})$ corresponds to the relative importance of a spatial information $(i,j)$ of a given feature map. This recalibration provides more importance to relevant spatial locations and ignores irrelevant ones. The architectural flow is shown in Fig.~\ref{fig:GA}(c). 

\noindent
\subsection{Spatial and Channel Squeeze \& Excitation Block (scSE)}
Finally, we introduce a combination of the above two SE blocks, which concurrently recalibrates the input $\mathbf{U}$ spatially and channel-wise. We obtain the concurrent spatial and channel SE, $\hat{\mathbf{U}}_{scSE}$, by element-wise addition of the channel and spatial excitation, $\hat{\mathbf{U}}_{scSE} = \hat{\mathbf{U}}_{cSE} + \hat{\mathbf{U}}_{sSE}$. A location $(i,j,c)$ of the input feature map $\mathbf{U}$ is given higher activation when it gets high importance from both, channel re-scaling and spatial re-scaling. This recalibration encourages the network to learn more meaningful feature maps, that are relevant both spatially and channel-wise. The architecture of the combined scSE block is illustrated in Fig.~\ref{fig:GA}(d).

\vspace{4mm}
\noindent
\textbf{Model Complexity:}
Let us consider an encoder/decoder block, with an output feature map of $C$ channels. Addition of a cSE block introduces $C^2$ new weights, while a sSE block introduces $C$ weights. So, the increase in model complexity of a F-CNN with $h$ encoder/decoder blocks is $\sum_{i=1}^h (C_i^2 + C_i)$, where $C_i$ is the number of output channels for the $i^{th}$ encoder/decoder block. To give a concrete example, the U-Net in our experiments has about $2.1 \times 10^6$ parameters. The scSE block adds $3.3 \times 10^4$ parameters, which is an approximate increase by 1.5\%. Hence, SE blocks only increase overall network complexity by a very small fraction.

\section{Experimental Results}
In this section, we conducted extensive experiments to explore the impact of our proposed modules. We chose three state-of-the-art F-CNN architectures, U-Net~\cite{Unet}, SD-Net~\cite{ecb2017} and Fully Convolutional DenseNet~\cite{densenet}. All of the networks have an encoder/decoder based architecture. The encoding and decoding paths consist of repeating blocks separated by down-sampling and up-sampling, respectively. We insert (i) channel-wise SE (cSE) blocks, (ii) spatial SE (sSE) blocks and (iii) concurrent spatial and channel-wise SE (scSE) blocks after every encoder and decoder block of the F-CNN architecture and compare against its vanilla version.

\noindent
\textbf{Datatsets: }
We use two datasets in our experiments.  (i) Firstly, we tackle the task of segmenting MRI T1 brain scans into 27 cortical and sub-cortical structures. We use the Multi-Atlas Labelling Challenge (MALC) dataset~\cite{malc}, which is a part of OASIS~\cite{oasis}, with 15 scans for training and 15 scans for testing consistent to the challenge instructions. The main challenge associated with the dataset are the limited training data with severe class imbalance between the target structures. Manual segmentations for MALC were provided by Neuromorphometrics, Inc.\footnote{http://Neuromorphometrics.com/} (ii) Secondly, we tackle the task of segmenting 10 organs on whole-body contrast enhanced CT (ceCT) scans. We use data from the Visceral dataset~\cite{visceral}. We train on 65 scans from the silver corpus, and test on 20 scans with manual annotations from the gold corpus. The silver corpus was automatically labeled by fusing the results of multiple algorithms, yielding noisy labels. The main challenge associated with the whole-body segmentation is the highly variable shape of the visceral organs and the capability to generalize when trained with noisy labels. We use Dice score for performance evaluation.

\noindent
\textbf{Model Learning:} In our experiments, all of the three F-CNN architectures had 4 encoder blocks, one bottleneck layer, 4 decoder blocks and a classification layer at the end. The logistic loss function was weighted with median frequency balancing~\cite{segnet} to compensate for the class imbalance. The learning rate was initially set to $0.01$ and decreased by one order after every 10 epochs. The momentum was set to $0.95$, weight decay constant to $10^{-4}$ and a mini batch size of $4$. Optimization was performed using stochastic gradient descent. Training was continued till validation loss converged. All the experiments were conducted on an NVIDIA Titan Xp GPU with 12GB RAM.

\begin{table}[t]
\caption{Mean and standard deviation of the global Dice scores for the different F-CNN models without and with cSE, sSE and scSE blocks on both datasets.}
\centering
\begin{tabular}{|p{0.94in}|P{0.89in}|P{0.89in}|P{0.89in}|P{0.89in}|}
    \hline
    & \multicolumn{4}{c|}{\textbf{MALC Dataset}} \\
     Networks & No SE Block & + cSE Block & + sSE Block & + scSE Block  \\
    \hline
    \textbf{DenseNets}\cite{densenet} & $0.842\pm0.058$ & $0.865\pm0.069$ & $0.876\pm0.061$ & $\mathbf{0.882}\pm0.063$  \\ 
    \textbf{SD-Net}\cite{ecb2017} & $0.771\pm0.150$ & $0.790\pm0.120$ & $0.860\pm0.062$ & $\mathbf{0.862}\pm0.082$   \\ 
    \textbf{U-Net}\cite{Unet} & $0.763\pm0.110$ & $0.825\pm0.063$ & $0.837\pm0.058$ & $\mathbf{0.843}\pm0.062$ \\ \hline
    & \multicolumn{4}{c|}{\textbf{Visceral Dataset}} \\
      Networks & No SE Block & + cSE Block & + sSE Block & + scSE Block  \\
    \hline
    \textbf{DenseNets}\cite{densenet} & $0.892\pm0.068$ & $0.903\pm0.058$ & $0.912\pm0.056$ & $\mathbf{0.918}\pm0.051$  \\ 
    \textbf{SD-Net}\cite{ecb2017} & $0.871\pm0.064$ & $0.892\pm0.065$ & $0.901\pm0.057$ & $\mathbf{0.907}\pm0.057$   \\ 
    \textbf{U-Net}\cite{Unet} & $0.857\pm0.106$ & $0.865\pm0.086$ & $0.872\pm0.080$ & $\mathbf{0.881}\pm0.082$ \\ \hline
  \end{tabular}

  \label{tab:res}
\end{table}

\noindent
\textbf{Quantitative Results: }
Table~\ref{tab:res} lists the mean Dice score on test data for both datasets. Results of the standard networks together with the addition of cSE, sSE and scSE blocks are reported. Comparing along the columns, we observe that inclusion of any SE block consistently provides a statistically significant ($p\le0.001$, Wilcoxon signed-rank) increase in Dice score in comparison to the normal version for all networks, in both  applications. We further observe that the spatial excitation yields a higher increase than the channel-wise excitation, which confirms our hypothesis that spatial excitation is more important for segmentation. Spatial and channel-wise SE yields the overall highest performance, with an increase of $4-8 \%$ Dice for brain segmentation and $2-3 \%$ Dice for whole-body segmentation compared to the standard network. Particularly for brain, the performance increase is striking, given the limited increase in model complexity. Comparing the results across network architectures, DenseNets yield the best performance. 

\begin{figure}[t]
\centering
\includegraphics[width=0.95\textwidth]{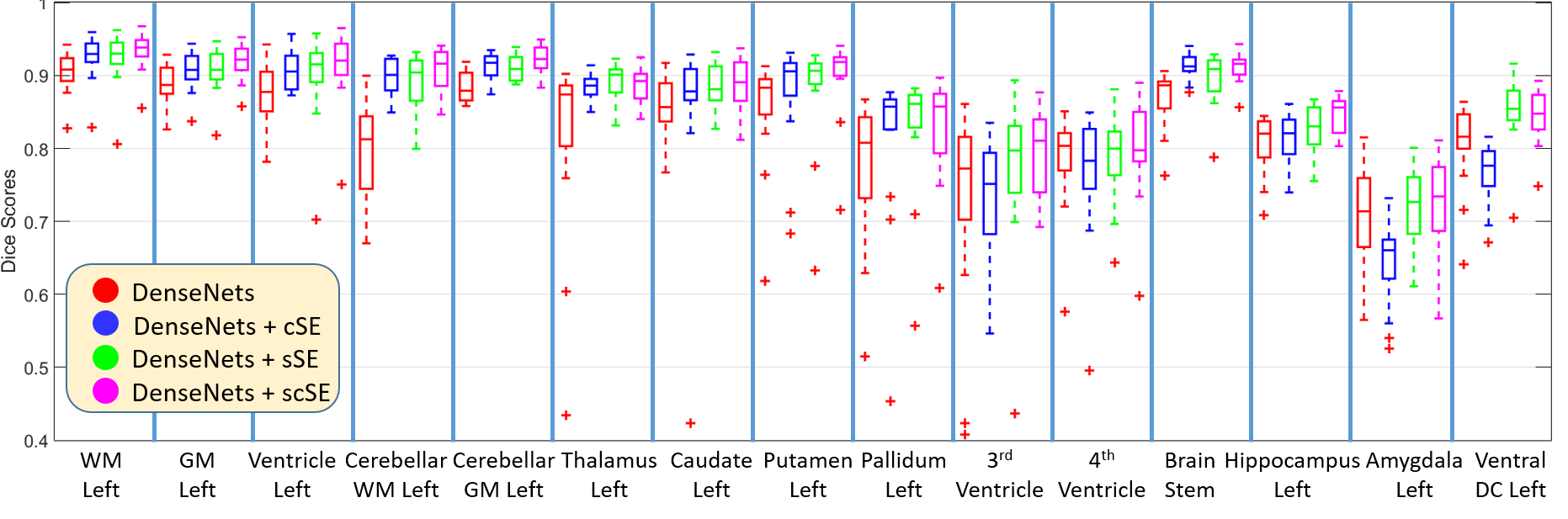}
\vspace{-2mm}
\caption{Boxplot of Dice scores for all brain structures on the left hemisphere (due to space constraints), using DenseNets on MALC dataset, without and with proposed cSE, sSE, scSE blocks. Grey and white matter are abbreviated as GM and WM, respectively.}
\label{fig:plotBrain}
\end{figure}

\begin{figure}[t]
\centering
\includegraphics[width=0.90\textwidth]{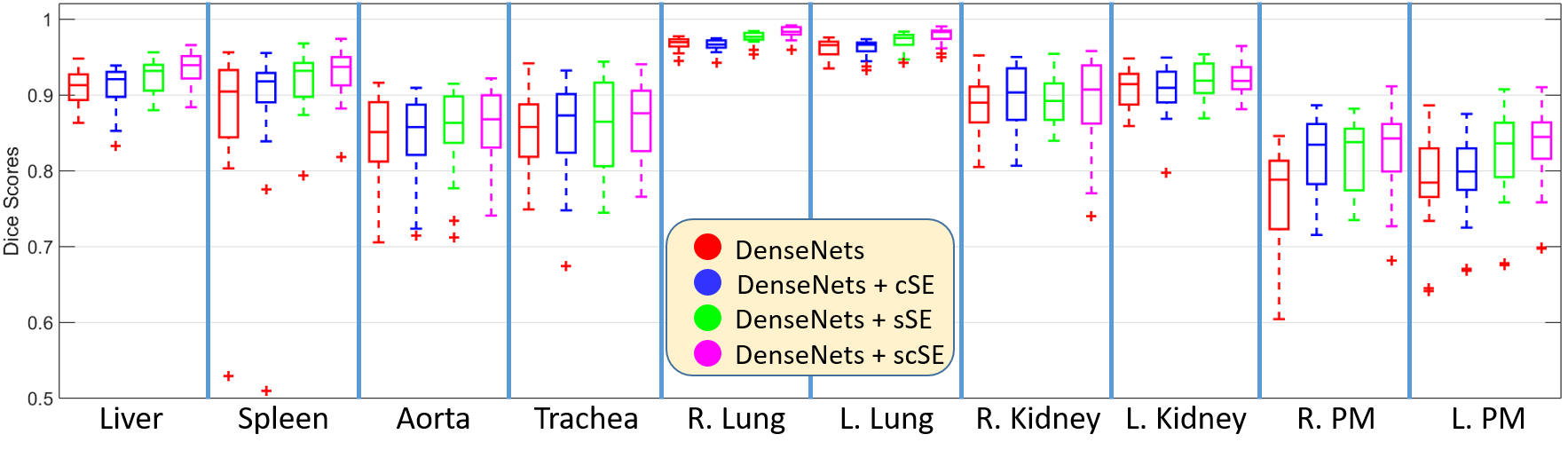}
\vspace{-2mm}
\caption{Structure-wise Dice performance of DenseNets on Visceral dataset, without and with proposed cSE, sSE, scSE blocks. Left and right are indicated as L. and R. Psoas major muscle is abbreviated as PM.}
\label{fig:plotBody}
\end{figure}

Fig.~\ref{fig:plotBrain} and Fig.~\ref{fig:plotBody} present structure-wise results for whole brain and whole body segmentation, respectively, for DenseNets. In Fig.~\ref{fig:plotBrain}, we observe that sSE and scSE outperform the normal model consistently for all the structures. The cSE model outperforms the normal model in most structures except some challenging structures like 3rd/4th ventricles, amygdala and ventral DC where its performance degrades. One possible explanation could be the small size of these structures, which might have got overlooked by only exciting the channels. The performance of sSE and scSE is very close. For whole body segmentation, in Fig.~\ref{fig:plotBody}, we observe a similar pattern.

\begin{figure}[t]
\centering
\includegraphics[width=0.9\textwidth]{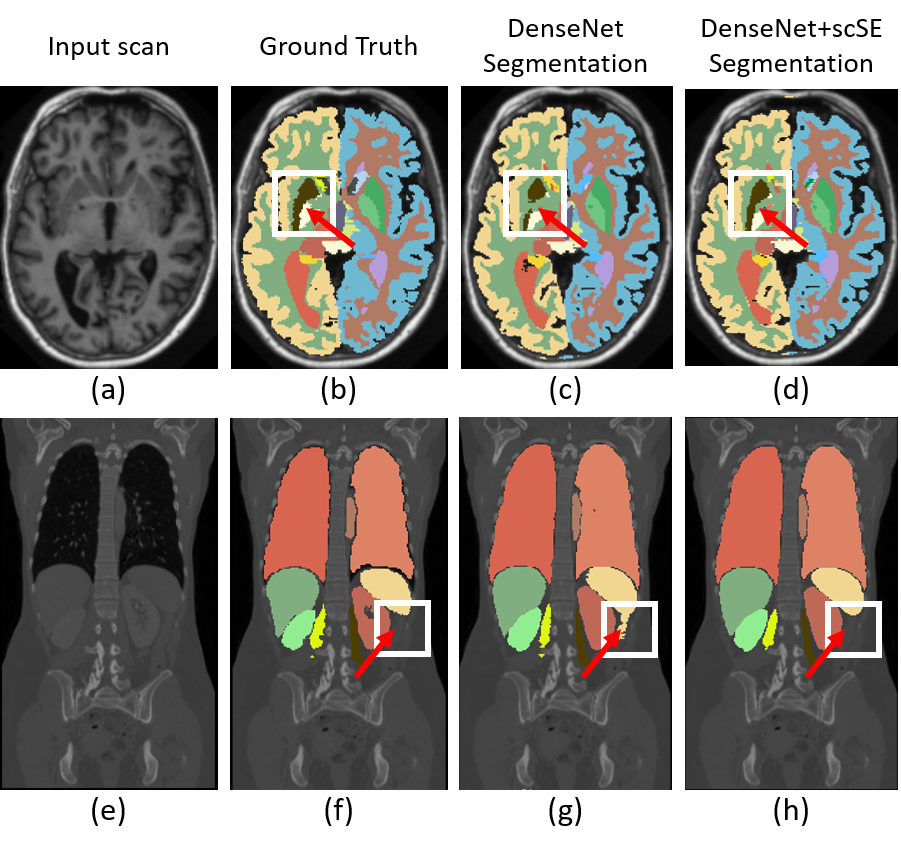}
\vspace{-2mm}
\caption{Input scan, ground truth annotations, DenseNet segmentation and DenseNet+scSE segmentation for both whole-brain MRI T1 (a-d) and whole-body ceCT (e-h) are shown.  ROIs are indicated by white box and red arrow highlighting regions where the scSE block improved the segmentation, for both applications.}
\label{fig:result}
\end{figure}

\noindent
\textbf{Qualitative Results: }
Fig.~\ref{fig:result}  presents segmentation results for MRI T1 brain scan in Fig.~\ref{fig:result}(a-d) and for Whole body ceCT scans in Fig.~\ref{fig:result}(e-h). We show the input scan, ground truth annotations, DenseNet segmentation along with our proposed DenseNet+scSE segmentation. We highlight ROIs with a white box and red arrow, to show regions where inclusion of scSE block improved the segmentation. For MRI brain scan segmentation, we indicate the structure left putamen, which is under segmented using  DenseNet (Fig.~\ref{fig:result}(c)), but the segmentation improves with the inclusion of the scSE block (Fig.~\ref{fig:result}(d)). For whole body ceCT, we indicate the  spleen, which is over segmented using DenseNet (Fig.~\ref{fig:result}(g)), and which is rectified with adding scSE block (Fig.~\ref{fig:result}(h)).

\vspace{-2mm}
\section{Conclusion}
\label{sec:conc}
\vspace{-2mm}
We proposed the integration of squeeze \& excitation  blocks within F-CNNs for image segmentation. Further, we introduced the \emph{spatial} squeeze \& excitation, which outperforms the previously proposed channel-wise squeeze \& excitation. We demonstrated that SE blocks yield a consistent improvement for three different F-CNN architectures and for two different segmentation applications. Hence, recalibration with SE blocks seems to be a fairly generic concept to boost performance in CNNs. Strikingly, the substantial increase in segmentation accuracy comes with a negligible increase in model complexity. With the seamless integration, we believe that squeeze \& excitation can be a crucial component for neural networks in many medical applications. 

\noindent
\textbf{Acknowledgement:} We thank the Bavarian State Ministry of Education, Science and the Arts in the framework of the Centre Digitisation.Bavaria (ZD.B) for funding and NVIDIA corporation for GPU donation.

\end{document}